\documentclass{article} 
\usepackage{iclr2024_conference,times}

\usepackage{amsmath,amsfonts,bm}

\def\eqref#1{equation~\ref{#1}}

\def\1{\bm{1}}

\DeclareMathAlphabet{\mathsfit}{\encodingdefault}{\sfdefault}{m}{sl}
\SetMathAlphabet{\mathsfit}{bold}{\encodingdefault}{\sfdefault}{bx}{n}

\usepackage{hyperref}
\usepackage{url}
\usepackage{amsmath,epsfig}
\usepackage[utf8]{inputenc} 
\usepackage{tabularray}
\usepackage{algorithmic}
\usepackage[ruled,vlined]{algorithm2e}

\title{Region-level Labels in Ice Charts can produce Pixel-level Segmentation for Sea Ice Types}

\author{
Muhammed Patel$^{1}$\thanks{Corresponding author. Email: \texttt{m32patel@uwaterloo.ca}} \quad Xinwei Chen$^{2}$ \quad Linlin Xu$^{1}$ \quad Yuhao Chen$^{1}$ \\ \quad \textbf{K Andrea Scott}$^{3}$ \quad \textbf{David A. Clausi}$^{1}$ \\
$^1$Vision and Image Processing Lab, Systems Design Engineering, University of Waterloo\\
$^2$School of Marine Science and Engineering, South China University of Technology\\
$^3$Department of Mechanical and Mechatronics Engineering University of Waterloo\\
}

\iclrfinalcopy

\begin{document}

\maketitle
\begin{abstract}
Fully supervised deep learning approaches have demonstrated impressive accuracy in sea ice classification, but their dependence on high-resolution labels presents a significant challenge due to the difficulty of obtaining such data. In response, our weakly supervised learning method provides a compelling alternative by utilizing lower-resolution regional labels from expert-annotated ice charts. This approach achieves exceptional pixel-level classification performance by introducing regional loss representations during training to measure the disparity between predicted and ice chart-derived sea ice type distributions. Leveraging the AI4Arctic Sea Ice Challenge Dataset, our method outperforms the fully supervised U-Net benchmark, the top solution of the AutoIce challenge, in both mapping resolution and class-wise accuracy, marking a significant advancement in automated operational sea ice mapping.

\end{abstract}

\section{Introduction}
The increasing impact of rapid climate change on polar sea ice has heightened the importance of monitoring sea ice types using operational Synthetic Aperture Radar (SAR) imagery. This monitoring is crucial for maritime safety, Arctic shipping route planning, and assimilation into forecasting models. SAR satellite observations are commonly used for sea ice mapping, providing wide coverage, day/night observation, high resolution, nearly all-weather capability, and a high revisit rate \citep{lyu2022meta}.

Distinct scattering properties in SAR imagery allow the differentiation of various ice types. Manual annotation of sea ice charts, a common practice for operational sea ice mapping, results in coarse spatial resolution and lacks pixel-level labels, motivating the development of automated methods. Current limitations in manual ice charting, including biases, representativity errors, and labeling uncertainties, underscore the need for more accurate and high-resolution automated sea ice mapping \cite{stokholm2022ai4seaice}. In recent years, deep learning (DL) models, such as convolutional neural networks (CNNs), have shown promise due to their ability to automatically extract features, with many related works being published (\cite{boulze2020classification,lyu2022eastern,chen2023sea}). However, fully-supervised CNNs require extensive labeled data, which is often limited in sea ice studies.

The AI4Arctic Sea Ice Challenge dataset by \cite{buus-hinkler2022ai4arctic} has emerged as a benchmark, providing a large-scale dataset for DL-based sea ice mapping. While this dataset addresses some limitations, challenges remain, such as uncertainties and errors introduced during label conversion.

The ice analyst annotates the SAR image using a polygon-based egg code. This egg code contains the concentration of each ice type within the polygon and lacks pixel-wise labels; therefore, the exact location of the label couldn't be known. However, training a semantic segmentation model requires pixel-wise labels. So, as a workaround, in the AI4Arctic dataset, only dominant ice-type polygons(partial concentration $>$ 65\%) are used and converted to pixel-wise labels. However, real-world SAR images contain lots of heterogeneous regions, and getting rid of them is not ideal, resulting in the loss of crucial data for training. 

To overcome the limitations of fully supervised learning, this paper introduces a novel weakly supervised learning method for per-pixel sea ice type classification directly from region-based ice chart labels, following the successful implementations of weakly supervised learning in remote sensing imagery \citep{xu2016weakly}. The methodology, experimental results, and conclusions are detailed in subsequent sections.

\section{Methodology}
\label{sec:method}
\subsection{Regional Label Generation from Ice Charts}
Despite the SIGRID-3 coding \citep{SIGRID-3} in ice charts offering over 10 classes of ice types, some are underrepresented, leading to label uncertainty and visual ambiguities between similar classes. To simplify the problem and enhance generalization, we aggregate SIGRID-3 codes into four classes: open water (ice-free), young ice, first-year ice, and multiyear ice, as described in Table \ref{SOD class table}. Since the ice characteristics in ice charts are provided for each manually drawn polygon, the regional label for each polygon can be represented by a 4-dimensional vector, indicating the partial fraction of a given ice type. More details about the simplified egg code can be found in Appendix \ref{Appendix:Simplified Eggcode}.
\begin{table}[h]
\centering
\caption{Description of the classes of sea ice types defined in this research.}
\footnotesize	
\begin{tabular}{|c|c|l|}
\hline
\begin{tabular}[c]{@{}c@{}}Entry \\ number\end{tabular} & \begin{tabular}[c]{@{}c@{}}Class\\ representation\end{tabular} & \multicolumn{1}{c|}{Description}                                                                                                               \\ \hline
0                                                       & Open water                                                     & Ice free regions identified in ice charts                                                                                                      \\ \hline
1                                                       & Young ice                                                      & Ice with thickness no more than 30 cm                                                                                                          \\ \hline
2                                                       & \begin{tabular}[c]{@{}l@{}} First year ice \\ (FYI) \end{tabular}                                                & \begin{tabular}[c]{@{}l@{}}Ice of no more than one year's growth \\ developing from young ice with \\ thickness larger than 30 cm\end{tabular} \\ \hline
3                                                       &\begin{tabular}[c]{@{}l@{}} Multiyear ice \\ (MYI)    \end{tabular}                                                 & Ice survived at least one melt season                                                                                                                \\ \hline

\end{tabular}
\label{SOD class table}
\end{table}
\begin{figure*}[h]
           \centering
        \includegraphics[scale = 0.40]{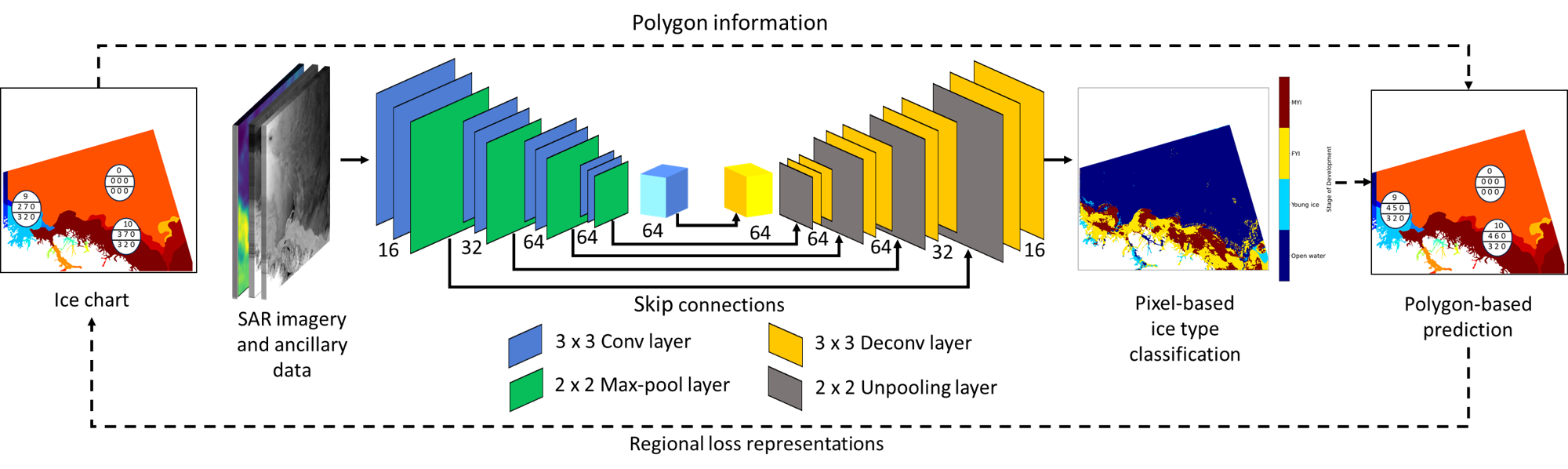}
        \caption{The architecture of the proposed pixel-level ice type mapping model based on a U-Net and the weakly supervised learning scheme (indicated by dash lines). The digit below each layer indicates the number of filters for each convolution/deconvolution layer.}
        \label{sod unet}  
\end{figure*}

\subsection{Network architecture}
We employ the U-Net architecture for semantic segmentation \citep{ronneberger2015u} as the foundation of our ice type classification model, as shown in Fig. \ref{sod unet}. The input channels include Sentinel-1 SAR imagery and corresponding multi-source ancillary data, as described in Appendix (Table \ref{input channel table}). Other experiment-related hyperparameters, such as input channels, batch size, learning rate, etc., are also outlined in the Appendix \ref{Appendix:Model HyperParameters}.

The U-Net structure in use comprises four encoder-decoder blocks. For the encoder, the number of filters (with $3\times 3$ kernel size) in the $1^{st}$ to $4^{th}$ layer is 16, 32, 64, and 64, respectively (and vice versa for the decoder). To predict pixel-based ice type classification, the outputs of the last decoder are processed through a $1\times1$ convolution layer with the number of filters(channels) equal to the number of classes.

\subsection{Weakly Supervised Learning via Regional Loss Representations}

Our proposed algorithm employs a weakly supervised learning approach, as depicted in Fig. \ref{sod unet}. It uses a U-Net architecture to generate a pixel-wise classification map that identifies different types of ice in an input image. During the training phase, we assume each image patch contains an arbitrary number of polygons that have been manually annotated on ice charts. For each of these polygons, we derive a prediction tensor from the U-Net's output. This prediction tensor, which is a 4-dimensional vector, represents the estimated partial concentrations of various ice classes within the polygon. We then compare this polygon-based prediction tensor against a ground truth tensor derived from the ice chart and calculate the loss. This comparison enables the model to learn from regional (polygon-based) representations, and produce pixel-by-pixel ice types predictions.

Mathematically, if the predicted class probabilities for each pixel in a polygon are denoted as
$\mathbf{y} = \{y_{1}, y_{2}, y_{3}, y_{4}\}$, then each element in the 4 dimensional polygon-based predicted tensor (denoted as $\mathbf{Y}$) can be calculated as


\begin{equation}
    Y_{i} = \frac{1}{N_{pixel}} \sum_{j=1}^{N_{pixel}} y_{i}^{j}, \quad i = 1,...,4,
    \label{polygon predicted vecotr calculation}
\end{equation}
where $N_{pixel}$ refers to the total number of pixels inside a polygon, and $y_{i}^{j}$ represents $y_{i}$ for the $j^{th}$ pixel. In this work, we use the cross-entropy (CE) loss, expressed as
\begin{equation}
\text{loss} = -\frac{1}{n} \sum_{i=1}^{n} Y_{i} \log(\hat{Y}_i),
\label{ce loss}
\end{equation}
where $n$ is the number of elements (classes) in each predicted/ground truth polygon-based tensor, and $Y_i$ and $\hat{Y}_i$ are the ground truth and predicted concentrations for the $i^{th}$ class from the polygon, respectively. Finally, the loss calculated from all the polygons in all image patches in a batch is aggregated as the total loss for backpropagation. By minimizing this total loss using optimization algorithms (e.g., stochastic gradient descent \cite{loshchilov2016sgdr}), the model effectively learns the distribution of different ice types in each polygon while accurately providing per-pixel classification. The pseudocode for this loss function can be found in Appendix \ref{Appendix: Psuedo code}
\section{Experimental Results}
\label{sec:results}
We perform our experiments on the AI4arctic dataset and follow the same train-test split as in the Autoice competition (\citep{stokholm2023autoice}) for direct comparison. Out of the 533 dataset files, 22 are reserved for testing, and cross-validation is performed on the training set. Each run involves randomly selecting 20 scenes for validation, with a total of 30 runs ensuring extensive scene coverage.

Given the absence of pixel-based ground truth, we numerically evaluate the model by computing R2 scores between polygon-based sea ice type labels and predictions representing partial concentrations. The R2 score for a specific class $R_{i}^{2}$ is calculated using the formula:
\begin{equation}
    R_{i}^{2}=1-\frac{\sum_{k=1}^{N_{poly}}(Y_{i}^{k}-\hat{Y}_{i}^{k})^{2}}{\sum_{k=1}^{N_{poly}}(Y_{i}^{k}-\bar{Y}_{i})^{2}},
    \label{r2_score}
\end{equation}
where $Y_{i}^{k}$ and $\hat{Y}_{i}^{k}$ are the ground truth and predicted partial concentration for the $i^{th}$ class in the $k^{th}$ polygon, $\bar{Y}_{i}$ is the mean value of $Y_{i}^{k}$ across all polygons, and $N_{poly}$ is the total number of polygons in the validation/testing set.

\begin{table*}[h]
\centering
\scriptsize
\caption{The class-wise R2 scores calculated between predictions and ice charts of each ice type using the benchmark U-Net and the proposed weakly supervised U-Net. The improvement of R2 scores for each class using the weakly supervised model are indicated by the percentages in the brackets.}
\label{r2 score table}
\begin{tblr}{
  cells = {c},
  cell{1}{1} = {r=2}{},
  cell{1}{2} = {c=4}{},
  cell{1}{6} = {c=4}{},
  vlines,
  hline{1,3-5} = {-}{},
  hline{2} = {2-9}{},
}
Models                                             & Cross validation      &                        &                       &                       & Testing               &                        &                       &                       \\
                                                   & Open water            & Young ice              & First year ice        & Multiyear ice         & Open water            & Young ice              & First year ice        & Multiyear ice         \\
{Benchmark U-Net trained \\ with per pixel labels} & 93.95\%               & 57.57\%                & 75.54\%               & 80.78\%               & 91.37\%               & 41.22\%                & 75.52\%               & 83.61\%               \\
\textbf{Weakly supervised U-Net}                   & {97.45\%\\ (\textbf{+4.00\%})} & {68.16\%\\ (\textbf{+10.59\%})} & {85.23\%\\ (\textbf{+9.69\%})} & {88.93\%\\ (\textbf{+8.17\%})} & {95.73\%\\ (\textbf{+4.36\%})} & {58.83\%\\ (\textbf{+17.61\%})} & {83.09\%\\ (\textbf{+7.57\%})} & {86.04\%\\ (\textbf{+2.43\%})} 
\end{tblr}
\end{table*}

We compare our weakly supervised U-Net with the winner of the AutoIce competition, referred to as benchmark U-Net (\citet{chen2024mmseaice}). We keep all the hyperparameters (\ref{Appendix:Model HyperParameters}) same and only replace the loss function and ground truth used. The benchmark model is trained with pixel-wise labels and cross-entropy loss. Table \ref{r2 score table} displays class-wise R2 scores computed between predictions and ice charts labels on validation and testing sets. The weakly supervised U-Net achieves higher R2 scores in all ice types. In the cross-validation sets, it increases R2 scores by around 4\%, 10.59\%, 9.69\%, and 8.17\% for open water, young ice, first-year ice, and multiyear ice, respectively. The same improvements are observed in the testing set: 4.36\%, 17.61\%, 7.57\%, and 2.43\%. The hypothesis for these improvements is that the benchmark U-Net only uses pixel-wise labels derived from dominant polygons and ignores mixed polygons. This 
gets rid of so many polygons, which usually contain minority classes (Young ice, First-year ice). On the other hand, our method is more aligned with the way the labels are defined in charts.
\begin{figure*}[h]
           \centering
        \includegraphics[scale = 0.28]{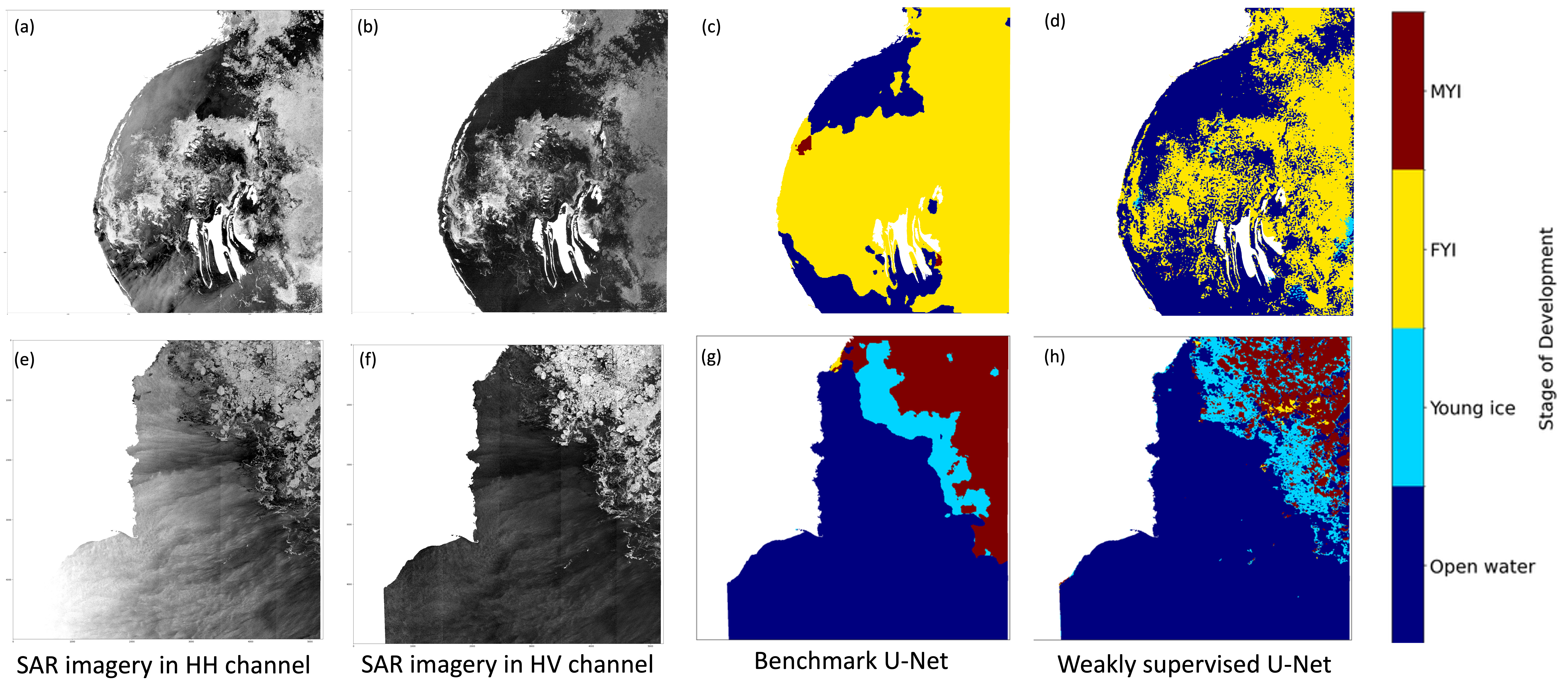}
        \vspace{-3mm}
        \caption{Visualization of results obtained from two SAR scenes. (a): SAR image in HH channel, file ID name: 20180707T113313\underline{\hspace{0.5em}}cis, collected on July $7^{th}$, 2018 in Hudson Bay (center latitude/longitude: $57.07^{\circ}N, 77.96^{\circ}W$); (b): The HV channel of the SAR image in (a); (c): Classification results of (a) from the benchmark U-Net model (dark blue: open water; light blue: young ice; yellow: first year ice; red: multiyear ice); (d): Classification results of (a) from the proposed weakly supervised U-Net; (e): SAR image in HH channel, file ID name: 20190926T152005\underline{\hspace{0.5em}}cis, collected on September $26^{th}$, 2019 in Western Arctic (center latitude/longitude: $72.63^{\circ}N, 126.61^{\circ}W$); (f): The HV channel of the SAR image in (e); (g): Classification results of (e) from the benchmark U-Net model; (h): Classification results of (e) from the proposed weakly supervised U-Net. The land area is masked in white.}
        \label{result visual3}  
 \end{figure*}

While numerical results offer insights into accuracy, visual analysis in Fig. \ref{result visual3} provides a more qualitative assessment. The weakly supervised model significantly enhances mapping resolution, particularly in identifying ice-water boundaries, polygon-based sea ice concentration, and partial concentrations of different ice types. This improvement is evident in challenging seasons, such as melt and freeze-up periods. For example, in the SAR scene on the top of Fig. \ref{result visual3} (melt period), the weakly supervised model is able to detect the detailed ice-water boundaries and ice with small floe size surrounded by water precisely, while the benchmark model can only detect ice or water-dominated areas with coarse resolution. For the SAR scene at the bottom obtained in freeze-up, pixels dominated by young ice (with relatively low backscatter intensities) and multiyear ice (with very bright pixel intensities) are well separated with significantly enhanced resolution using the weakly supervised model. Additional visualization with zoom-in example can be found in Appendix \ref{Appendix: Extra Visualization}\par

\section{Conclusion}
We introduce an automated pipeline for sea ice type classification using SAR imagery and ancillary data, leveraging a weakly supervised U-Net. This approach holds potential for integration into operational sea ice mapping for tactical navigation. By enabling precise prediction of sea ice type distributions with high resolution, our method eliminates the need for per-pixel labels during model training. To address the challenge of limited high-resolution ground truths, our classification model employs weakly supervised learning techniques, incorporating regional labels and loss representations. This allows the utilization of region labels derived from ice charts with coarse resolution directly for model training. The model is trained and validated using the recently released AI4Arctic Sea Ice Challenge Dataset. Results demonstrate strong agreement between predictions and ice-type distributions in ice charts, validated through visual interpretation. In future works, we'll aim to test our method on other datasets and state-of-the-art segmentation models.

\section{Acknowledgement}
The authors acknowledge the providers of the AI4Arctic dataset and the organizers of the AutoIce competition. This work was supported in part by Environment and Climate Change Canada (ECCC) and the Natural Sciences and Engineering Research Council (NSERC) of Canada. This research was enabled in part by support provided by Calcul Québec’s (calculquebec.ca) and the Digital Research Alliance of Canada (alliancecan.ca).

\newpage

\bibliography{iclr2024_conference}

\begin{thebibliography}{13}
\providecommand{\natexlab}[1]{#1}
\providecommand{\url}[1]{\texttt{#1}}
\expandafter\ifx\csname urlstyle\endcsname\relax
  \providecommand{\doi}[1]{doi: #1}\else
  \providecommand{\doi}{doi: \begingroup \urlstyle{rm}\Url}\fi

\bibitem[SIG()]{SIGRID-3}
{IICWG: SIGRID-3: a vector archive format for sea ice charts}.
\newblock \url{https://library.wmo.int/records/item/37171-sigrid-3-a-vector-archive-format-for-sea-ice-charts}.
\newblock Accessed: 2023-10-20.

\bibitem[Boulze et~al.(2020)Boulze, Korosov, and Brajard]{boulze2020classification}
Hugo Boulze, Anton Korosov, and Julien Brajard.
\newblock Classification of sea ice types in {Sentinel-1 SAR} data using convolutional neural networks.
\newblock \emph{Remote Sens.}, 12\penalty0 (13):\penalty0 2165, 2020.
\newblock \doi{10.3390/rs12132165}.

\bibitem[Buus-Hinkler et~al.(2022)Buus-Hinkler, Wulf, Stokholm, Korosov, Saldo, Pedersen, and et~al.]{buus-hinkler2022ai4arctic}
Jørgen Buus-Hinkler, Tore Wulf, Andreas~Rønne Stokholm, Anton Korosov, Roberto Saldo, Leif~Toudal Pedersen, and et~al.
\newblock {AI4Arctic Sea Ice Challenge Dataset}.
\newblock {Technical University of Denmark. Collection}, 2022.
\newblock URL \url{https://doi.org/10.11583/DTU.c.6244065.v2}.

\bibitem[Chen et~al.(2023)Chen, Scott, Jiang, Fang, Xu, and Clausi]{chen2023sea}
Xinwei Chen, K~Andrea Scott, Mingzhe Jiang, Yuan Fang, Linlin Xu, and David~A Clausi.
\newblock Sea ice classification with dual-polarized {SAR} imagery: A hierarchical pipeline.
\newblock In \emph{Proc. IEEE/CVF Winter Conf. Appl. Comput. Vis.}, pp.\  224--232, 2023.

\bibitem[Chen et~al.(2024)Chen, Patel, Pena~Cantu, Park, Noa~Turnes, Xu, Scott, and Clausi]{chen2024mmseaice}
Xinwei Chen, Muhammed Patel, Fernando~J Pena~Cantu, Jinman Park, Javier Noa~Turnes, Linlin Xu, K~Andrea Scott, and David~A Clausi.
\newblock Mmseaice: a collection of techniques for improving sea ice mapping with a multi-task model.
\newblock \emph{The Cryosphere}, 18\penalty0 (4):\penalty0 1621--1632, 2024.

\bibitem[{Joint {WMO-IOC} Technical Commission for Oceanography and Marine Meteorology}(2014)]{WMO_ice_chart_colour_code}
{Joint {WMO-IOC} Technical Commission for Oceanography and Marine Meteorology}.
\newblock Ice chart colour code standard, version 1.0, 2014.
\newblock Technical Report (WMO TD: 1215b), World Meteorological Organization \& Intergovernmental Oceanographic Commission, Geneva, Switzerland, 2014.

\bibitem[Loshchilov \& Hutter(2016)Loshchilov and Hutter]{loshchilov2016sgdr}
Ilya Loshchilov and Frank Hutter.
\newblock {SGDR}: Stochastic gradient descent with warm restarts.
\newblock \emph{arXiv preprint arXiv:1608.03983}, 2016.
\newblock URL \url{https://arxiv.org/abs/1608.03983}.

\bibitem[Lyu et~al.(2022{\natexlab{a}})Lyu, Huang, and Mahdianpari]{lyu2022eastern}
Hangyu Lyu, Weimin Huang, and Masoud Mahdianpari.
\newblock {E}astern arctic sea ice sensing: First results from the {RADARSAT Constellation Mission} data.
\newblock \emph{Remote Sens.}, 14\penalty0 (5):\penalty0 1165, 2022{\natexlab{a}}.

\bibitem[Lyu et~al.(2022{\natexlab{b}})Lyu, Huang, and Mahdianpari]{lyu2022meta}
Hangyu Lyu, Weimin Huang, and Masoud Mahdianpari.
\newblock A meta-analysis of sea ice monitoring using spaceborne polarimetric {SAR}: Advances in the last decade.
\newblock \emph{IEEE J. Sel. Top. Appl. Earth Obs. Remote Sens.}, 15:\penalty0 6158--6179, 2022{\natexlab{b}}.

\bibitem[Ronneberger et~al.(2015)Ronneberger, Fischer, and Brox]{ronneberger2015u}
Olaf Ronneberger, Philipp Fischer, and Thomas Brox.
\newblock {U-Net}: {Convolutional} networks for biomedical image segmentation.
\newblock In \emph{Intl. Conf. Med. Image Comput. Comput. Assist. Interv.}, pp.\  234--241. Springer, 2015.
\newblock \doi{10.1007/978-3-319-24574-4_28}.

\bibitem[Stokholm et~al.(2022)Stokholm, Wulf, Kucik, Saldo, Buus-Hinkler, and Hvidegaard]{stokholm2022ai4seaice}
Andreas Stokholm, Tore Wulf, Andrzej Kucik, Roberto Saldo, J{\o}rgen Buus-Hinkler, and Sine~Munk Hvidegaard.
\newblock {AI4SeaIce}: Toward solving ambiguous {SAR} textures in convolutional neural networks for automatic sea ice concentration charting.
\newblock \emph{IEEE Trans. Geosci. Remote Sens.}, 60:\penalty0 1--13, 2022.

\bibitem[Stokholm et~al.(2023)Stokholm, Buus-Hinkler, Wulf, Korosov, Saldo, Arthurs, Solberg, Long{\'e}p{\'e}, and Kreiner]{stokholm2023autoice}
Andreas Stokholm, J{\o}rgen Buus-Hinkler, Tore Wulf, Anton Korosov, Roberto Saldo, David Arthurs, Rune Solberg, Nicolas Long{\'e}p{\'e}, and Matilde Kreiner.
\newblock The {AutoICE} competition: Automatically mapping sea ice in the {A}rctic.
\newblock Technical report, Copernicus Meetings, 2023.

\bibitem[Xu et~al.(2016)Xu, Clausi, Li, and Wong]{xu2016weakly}
Linlin Xu, David~A Clausi, Fan Li, and Alexander Wong.
\newblock Weakly supervised classification of remotely sensed imagery using label constraint and edge penalty.
\newblock \emph{IEEE Trans. Geosci. Remote Sens.}, 55\penalty0 (3):\penalty0 1424--1436, 2016.

\end{thebibliography}
\bibliographystyle{iclr2024_conference}

\newpage
\appendix

\section{Appendix}

\subsection{Model HyperParameters}
\label{Appendix:Model HyperParameters}

\begin{table}[h]
\centering
\footnotesize
\caption{Specifications of training the proposed model.}
\begin{tabular}{|c|c|}
\hline
Optimizer                              & \begin{tabular}[c]{@{}c@{}}Stochastic gradient descent \\ with momentum (SGDM)\end{tabular} \\ \hline
Learning rate                          & 0.001                                                                                       \\ \hline
Weight decay                           & 0.01                                                                                        \\ \hline
Scheduler                              & Cosine Annealing                                                                            \\ \hline
Batch size                             & 16                                                                                          \\ \hline
Number of iterations per epoch         & 500                                                                                         \\ \hline
Total epoch                            & 50                                                                                          \\ \hline
Number of epochs for the first restart & 50                                                                                          \\ \hline
Image Downscaling ratio                      & 10                                                                                          \\ \hline
Patch size                             & 256                                                                                      \\ \hline
\end{tabular}
\label{training_table}
\end{table}

\begin{table}[h]
\centering
\footnotesize
\caption{The input channels of the proposed model.}
\begin{tabular}{|c|c|l|}
\hline
\begin{tabular}[c]{@{}c@{}}Input \\ type\end{tabular}        & \begin{tabular}[c]{@{}c@{}}Number of\\ channels\end{tabular} & \multicolumn{1}{c|}{Description}                                                                                                         \\ \hline
SAR imagery                                                     & 2                                                         & \begin{tabular}[c]{@{}l@{}}Dual polarized Sentinel-1 SAR images in \\ HH and  HV channels\end{tabular}                                              \\ \hline
\begin{tabular}[c]{@{}c@{}}Passive\\  microwave\end{tabular}    & 2                                                         & \begin{tabular}[c]{@{}l@{}}AMSR2 brightness temperature data in\\ 36.5 GHz in both horizontal and \\ vertical polarizations\end{tabular} \\ \hline
\begin{tabular}[c]{@{}c@{}}Spatial\\  information\end{tabular}  & 2                                                         & \begin{tabular}[c]{@{}l@{}}The latitude and logitude of each pixel \\ in the SAR imagery\end{tabular}                                    \\ \hline
\begin{tabular}[c]{@{}c@{}}Temporal\\  information\end{tabular} & 1                                                         & \begin{tabular}[c]{@{}l@{}}The acquisition month of the SAR \\ imagery\end{tabular}                                                                                                   \\ \hline
\end{tabular}
\label{input channel table}
\end{table} 

\subsection{Simplified Eggcode}
\begin{figure*}[h]
           \centering
        \includegraphics[scale = 0.45]{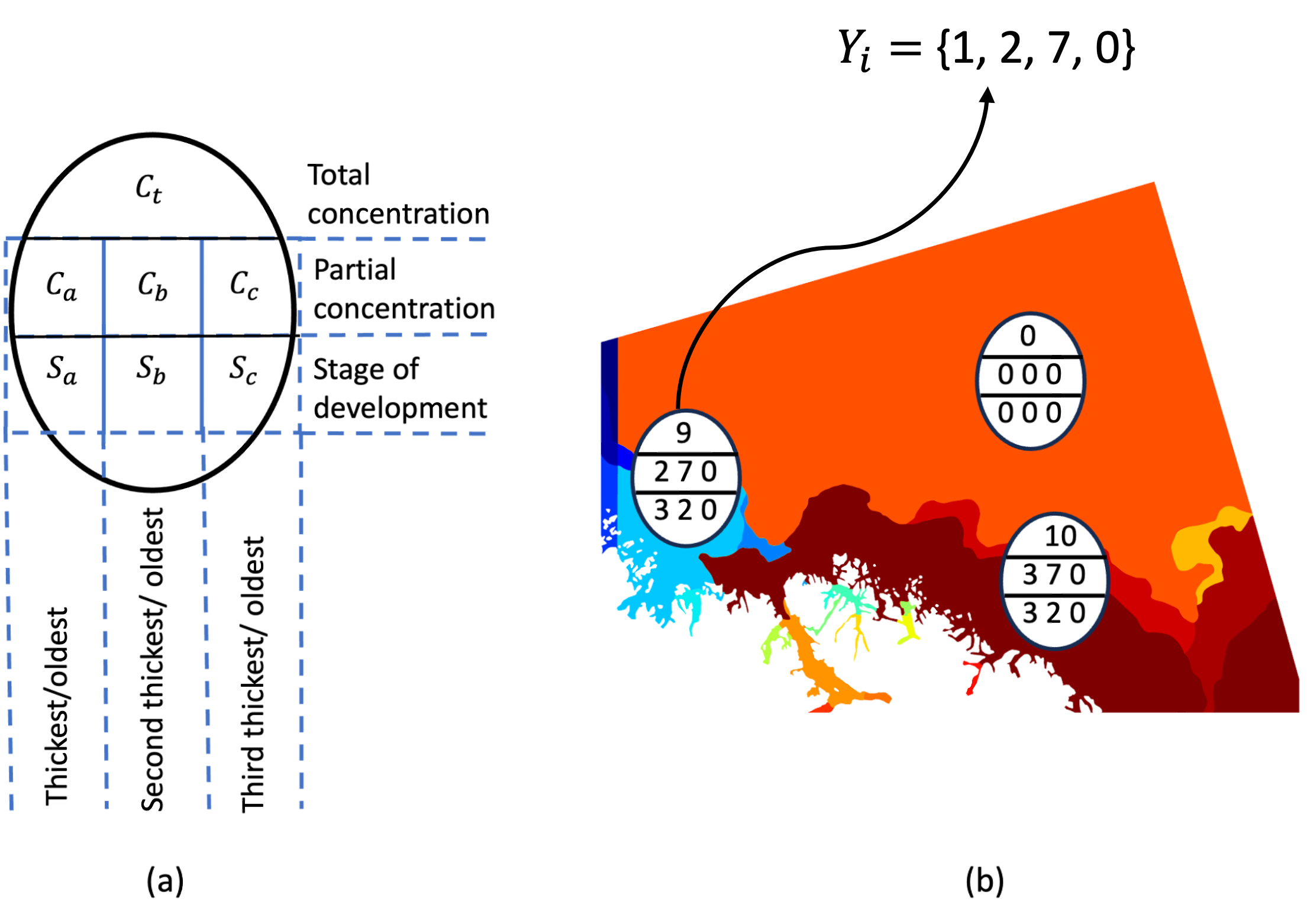}
        \vspace{-3mm}
        \caption{(a) Overview of the ice egg code parameters in the
simplified egg code produced from the original ice charts.
Concentration values range from 0 (empty) to 10 (fully-
covered) with each increase of 1 representing a 10 percent
concentration increase. (b) An example Ice chart showing eggcode and conversion of eggcode to ground truth tensor.}
        \label{simplified_eggcode} 
 \end{figure*}

An overview of the ice egg code variables in the simplified egg code is presented in Fig. \ref{simplified_eggcode} (a). Compared to the original egg code in WMO standard \citep{WMO_ice_chart_colour_code}, the information concerning the form of ice (floe size) is omitted. The total concentration, $C_{t}$ refers to the total SIC (Sea Ice concentration) of the polygon. As $C_{a}$, $C_{b}$, and $C_{c}$ correspond to the partial concentrations of the top 3 thickest ice types, there is a relation that $C_{a}+C_{b}+C_{c}=C_{t}$. On the other hand, $S_{a}$, $S_{b}$, and $S_{c}$ are represented by the entry numbers in Table \ref{SOD class table}, so for e.g, in the most left egg code in Fig. \ref{simplified_eggcode}(b), the value 9 represents $C_{t}$, values 2, 7, 0 denotes partial concentration $C_{a}$, $C_{b}$, $C_{c}$ respectively. Similarly, 3, 2, 0 represents the class $S_{a}$, $S_{b}$, $S_{c}$. And the final ground-truth 4 dimensional vector would be $Y_{i} = \{1, 2, 7, 0\}$ where the 1st element represents the concentration of water which is nothing but 1 - $C_{t}$.
\label{Appendix:Simplified Eggcode}

\subsection{Loss function: Psuedocode}
\label{Appendix: Psuedo code}
\begin{algorithm}[H]
\caption{Polygon-based Loss Calculation Pseudocode based on Python}
\KwData{polygon\_labels, pixel\_predictions, land\_masks, polygon\_maps}
\KwResult{total\_loss}
\SetAlgoNlRelativeSize{-1}
total\_loss = 0 \\
batch\_size = pixel\_predictions.shape[0] 

\For{img\_id in range(batch\_size)}{
polygon\_ids = np.unique(polygon\_maps[img\_id]);

\For{polygon\_id in polygon\_ids}{

\tcp{ get the label tensor for a certain polygon (polygon\_id)}
polygon\_label = polygon\_labels[img\_id][polygon\_id] \\

\tcp{ exclude the polygons that are masked out from loss calculation}
\If{polygon\_label == 255}{loss = 0 \\ total\_loss+=loss\\ \textbf{continue}\;}

\tcp{ create a mask to mask out pixels outside the current polygon}
poly\_mask = (polygon\_icecharts[img\_id] == poly\_id) \\

\tcp{ create final mask by removing land area from the poly\_mask}
final\_mask = poly\_mask $\cap$ land\_masks[img\_id]

\tcp{ pixel-based SOD predictions within the current polygon}
pixel\_AOI = pixel\_predictions[img\_id][final\_mask] \\

\tcp{ aggregate and average over pixel-based predictions to generate polygon-based prediction}
polygon\_prediction = pixel\_AOI.sum(0)/pixel\_AOI[0].numel() \\

\tcp{ calculate Cross Entropy loss between label tensor and predicted tensor}
loss = Cross\_Entropy(polygon\_prediction, polygon\_label) \\
total\_loss +=loss}
}
\Return{total\_loss}
\label{algorithm plot}
\end{algorithm}

\subsection{Additional Visualization}
\label{Appendix: Extra Visualization}
\begin{figure}[h!]
           \centering
        \includegraphics[scale = 0.35]{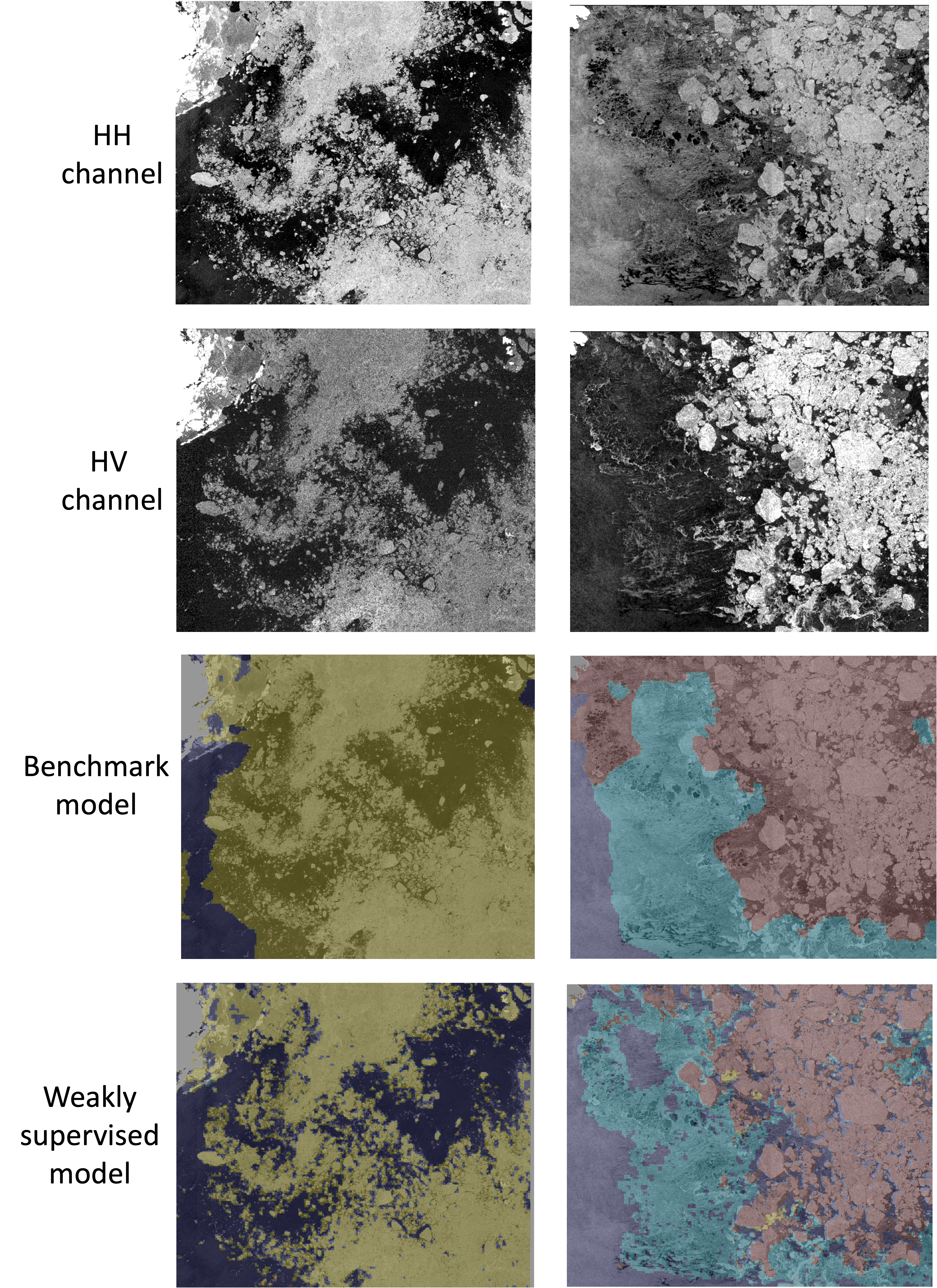}
        \caption{Zoom-in examples of regions cropped from SAR scenes and corresponding predictions in Fig. \ref{result visual3}. First column: a region located within a polygon in Fig. \ref{result visual3} dominated by first-year ice. Second column: a region located within a polygon in Fig. \ref{result visual3} consisting of both young ice and multiyear ice. First row: regions in HH channel; Second row: regions in HV channel; Third row: overlay of HH channel and predictions from benchmark model, with the color bar provided in Fig. \ref{result visual3}; Fourth row: overlay of HH channel and predictions from weakly supervised model.}
        \label{zoom in visual}  
 \end{figure}

The zoom-in examples from Fig. \ref{result visual3} are provided in Fig \ref{zoom in visual}. In the first column of Fig \ref{zoom in visual}, the weakly supervised model is able to detect the detailed ice-water boundaries and ice with small floe size surrounded by water precisely, while the benchmark model can only detect ice or water-dominated areas with coarse resolution. In the second column of Fig. \ref{zoom in visual}, pixels dominated by young ice (with relatively low backscatter intensities) and multiyear ice (with very bright pixel intensities) are well separated with each other using the weakly supervised model.

\end{document}